%% file: main.tex
\documentclass[10pt,twocolumn,letterpaper]{article}

\usepackage[pagenumbers]{cvpr} 

\usepackage{graphicx}
\usepackage{amsmath}
\usepackage{amssymb}
\usepackage{booktabs}
\usepackage{flushend}

\usepackage[pagebackref,breaklinks,colorlinks]{hyperref}

\usepackage[capitalize]{cleveref}
\usepackage{mathtools}
\crefname{section}{Sec.}{Secs.}
\Crefname{section}{Section}{Sections}
\Crefname{table}{Table}{Tables}
\crefname{table}{Tab.}{Tabs.}

\def\cvprPaperID{4095} 
\def\confName{CVPR}
\def\confYear{2023}

\begin{document}

    \title{Pretrained Diffusion Models for Unified Human Motion Synthesis}

    \author{Jianxin Ma, Shuai Bai, Chang Zhou\\
    DAMO Academy, Alibaba Group\\
    {\tt\small \{jason.mjx, baishuai.bs, ericzhou.zc\}@alibaba-inc.org}
    }
    \maketitle

    \begin{abstract}
        Generative modeling of human motion has broad applications in computer animation, virtual reality, and robotics.
        Conventional approaches develop separate models for different motion synthesis tasks, and typically use a model of a small size to avoid overfitting the scarce data available in each setting.
        It remains an open question whether developing a single unified model is feasible, which may 1) benefit the acquirement of novel skills by combining skills learned from multiple tasks, and 2) help in increasing the model capacity without overfitting by combining multiple data sources.
        Unification is challenging because 1) it involves diverse control signals as well as targets of varying granularity, and 2) motion datasets may use different skeletons and default poses.
        In this paper, we present MoFusion, a framework for unified motion synthesis.
        MoFusion employs a Transformer backbone to ease the inclusion of diverse control signals via cross attention, and pretrains the backbone as a diffusion model to support multi-granularity synthesis ranging from motion completion of a body part to whole-body motion generation.
        It uses a learnable adapter to accommodate the differences between the default skeletons used by the pretraining and the fine-tuning data.
        Empirical results show that pretraining is vital for scaling the model size without overfitting, and demonstrate MoFusion's potential in various tasks, e.g., text-to-motion, motion completion, and zero-shot mixing of multiple control signals.
        Project page: \url{https://ofa-sys.github.io/MoFusion/}.
    \end{abstract}


    \input{introduction.tex}
    \input{method.tex}
    \input{results.tex}
    \input{related.tex}
    \input{conclusion.tex}


    {\small
    \bibliographystyle{ieee_fullname}
    \bibliography{main}
    }

\end{document}

%% file: introduction.tex
\section{Introduction}
\label{sec:intro}

Learning generative models to automate the synthesis or completion of human motion clips has been a promising research direction, with broad applications in various industries such as computer animation of human characters, full-body avatar in virtual reality, and humanoid robotics.

Abundant tasks fall under the category of human motion synthesis.
For example, \emph{motion in-betweening}~\cite{ubi-inbetween,MotionInfilling,duan2021unify} is a task that involves generating intermediate frames between two keyframes, while \emph{text-to-motion}~\cite{ghosh2021-t2m,petrovich22temos,lin-vigil18,lang2pose} and \emph{music-to-dance}~\cite{li20-dance,siyao2022bailando,li2021learn,wenlinzhuang-dance,dance-revol} are two recent tasks that aim to generate motion clips pertinent to a text description or a piece of music, respectively.
There is also the task of \emph{inverse kinematics}~\cite{oreshkin2022protores}, which tries to find a set of joint parameters subject to some constraints while maintaining natural poses, e.g., how the joints should rotate such that the feet are placed at specified coordinates.
However, the existing approaches to motion synthesis typically investigate these tasks in an isolated manner, where independent models are developed for each task and each dataset.
And it is often not straightforward how a model proposed for one task can be easily reused by another task.

We believe that unifying motion synthesis tasks with one single model is of great value.
First, unification paves the way for the acquirement of novel skills by allowing the model to combine skills learned from multiple tasks, possibly in a zero-shot manner.
For instance, a unified model for both text-to-motion and music-to-dance may possess the ability to synthesize motion relevant to both a text and a music piece simultaneously, even if there is no training sample for text-music co-control.
Second, unification brings a larger corpus of data for training, since datasets from multiple tasks can be combined.
Unlike texts or images, motion data are scarce by nature, because they mostly come from either manually refined motion capture data or artist created animation clips.
The larger corpus make increasing the model size without overfitting feasible, and large model sizes are widely believed to correlate with better performance if given sufficient data~\cite{scaling-law}.

Designing a unified model for human motion synthesis poses challenges.
On one hand, the diverse motion synthesis tasks not only have control signals of different modalities such as texts, audio, and images, but also have prediction targets of different granularity.
For example, one task may ask the model to synthesis whole-body motion from scratch according to a piece of music, whereas another task may expect the model to modify the movement of a small body part in certain frames such that it fits a text description while keeping the movement of the other parts mostly unaltered.
On the other hand, motion datasets are notorious for not following the same convention, due to the specific needs of their creators.
In particular, each dataset can use a skeleton with a different number of joints, different joint indices, as well as a different default pose such as the T-pose, the A-pose, or any other poses.

In this paper, we present MoFusion, a framework that unifies various motion synthesis tasks while tackling the aforementioned challenges.
MoFusion views the process of human motion synthesis as a de-noising diffusion probabilistic model (DDPM)~\cite{ho-ddpm}, which provides great flexibility in supporting tasks of different granularity.
Whole-body motion is generated by fully de-noising a set of white noise frames, while altering a body part or drawing intermediate frames can be implemented by de-nosing only the region of interest and filling the other regions with known values.
We use the Transformer~\cite{attention} as the backbone of MoFusion to ease incorporating control signals of various modalities via cross attention, and we encode the signals using off-the-shelf encoders, e.g., a pretrained language model for texts.

MoFusion pretrains the motion decoder of the Transformer architecture using the $\mathbf{x}_0$-prediction variant~\cite{dalle2,diffusionlm} of the diffusion loss on a large corpus of motion data collected from multiple data sources.
The pretraining data are heuristically retargeted to a skeleton akin to the one defined by SMPL-H~\cite{SMPL,MANO}.
The pretrained model is then fine-tuned on downstream tasks,
where we wrap the motion decoder with a lightweight skeleton-aware adapter in case the dataset uses a skeleton different from the one used for pretraining.
The proposed skeleton adapter comprises trainable parameters for aligning joints and compensating for the differences in the default poses automatically.

The ability to perform novel tasks by combining learned skills naturally emerges after multitask pretraining.
For example, MoFusion can synthesize motion that simultaneously fits a text and a music piece, by alternating between the text encoder and the music encoder during the de-noising process.
This trick, which we name \emph{alternating control}, though still needs text-to-motion and music-to-dance data, does not require a training sample to contain both text and music labels.
It leverages the iterative nature of diffusion~\cite{ho-ddpm} and the flexibility of cross attention~\cite{attention}.

We highlight our core contributions as follows:
\begin{itemize}
    \item We propose MoFusion, a unified framework built upon a Transformer~\cite{attention} backbone pretrained as a diffusion model~\cite{ho-ddpm}, which achieves state-of-art performance on text-to-motion synthesis and motion in-betweening.
    \item We find that the diffusion model's de-noising pipeline supports \emph{alternating control}, a trick for mixing multiple control signals such as multiple texts and music clips, while needing no multi-control training sample.
    \item We show that the diffusion model is suitable for synthesis problems of different granularity, ranging from generating whole-body motions from scratch to completing the movement of a body part in certain frames.
    \item We demonstrate that pretraining is critical for realizing the performance gain of a larger-sized diffusion model.
    Severe overfitting occurs if the model is not pretrained.
    \item We design a learnable adapter to automatically bridge the gap between the different default skeletons used by the pretraining and the fine-tuning data.
\end{itemize}

\begin{figure*}
    \centering
    \begin{subfigure}[t]{0.28\linewidth}
        \centering
        \includegraphics[width=0.95\linewidth]{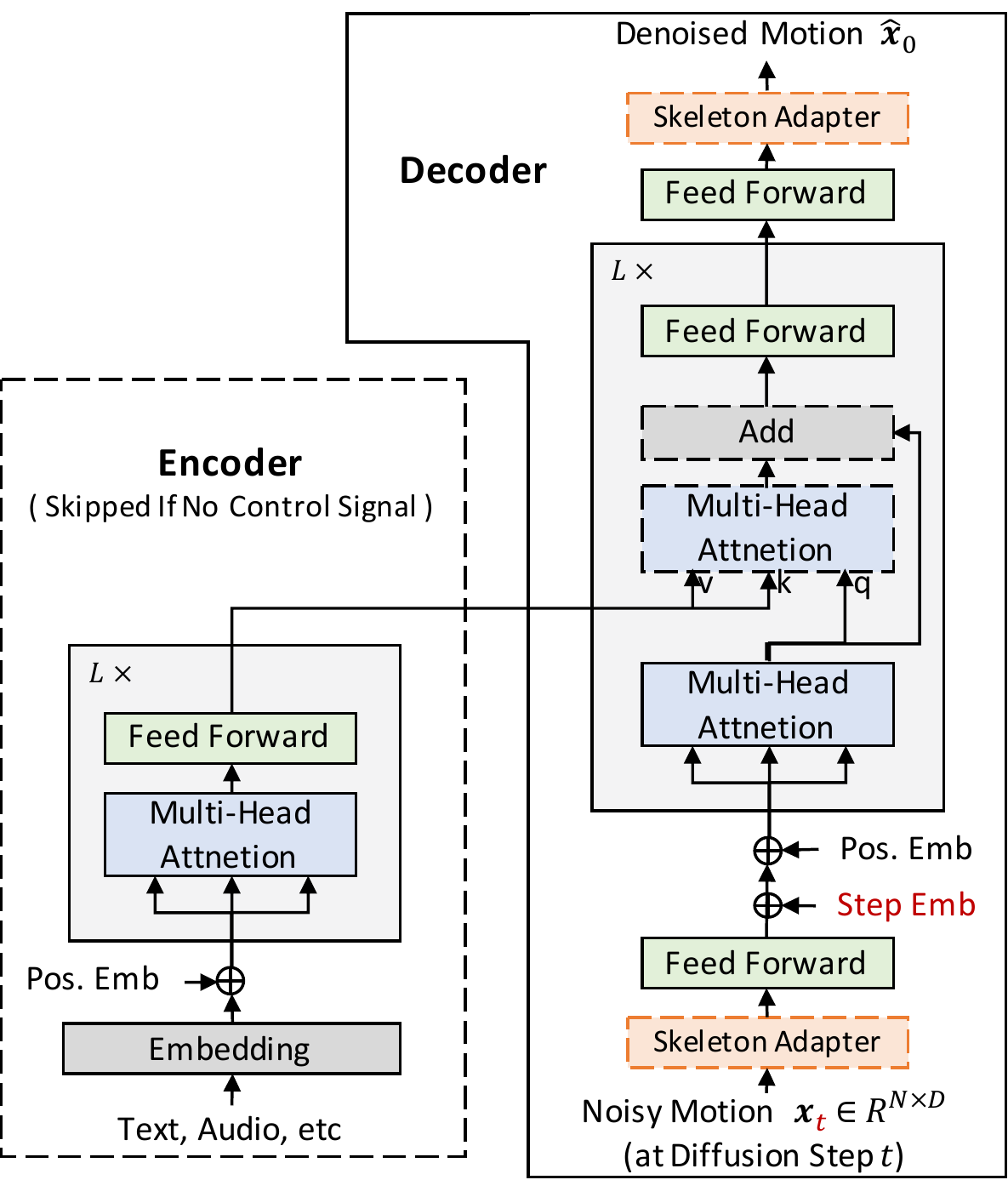}
        \caption{The overall architecture.
        An embedding corresponding to the current diffusion step $t$ is added to the token embeddings.
        }
        \label{fig:main-transformer}
    \end{subfigure}
    \hfill
    \begin{subfigure}[t]{0.70\linewidth}
        \centering
        \includegraphics[width=\linewidth]{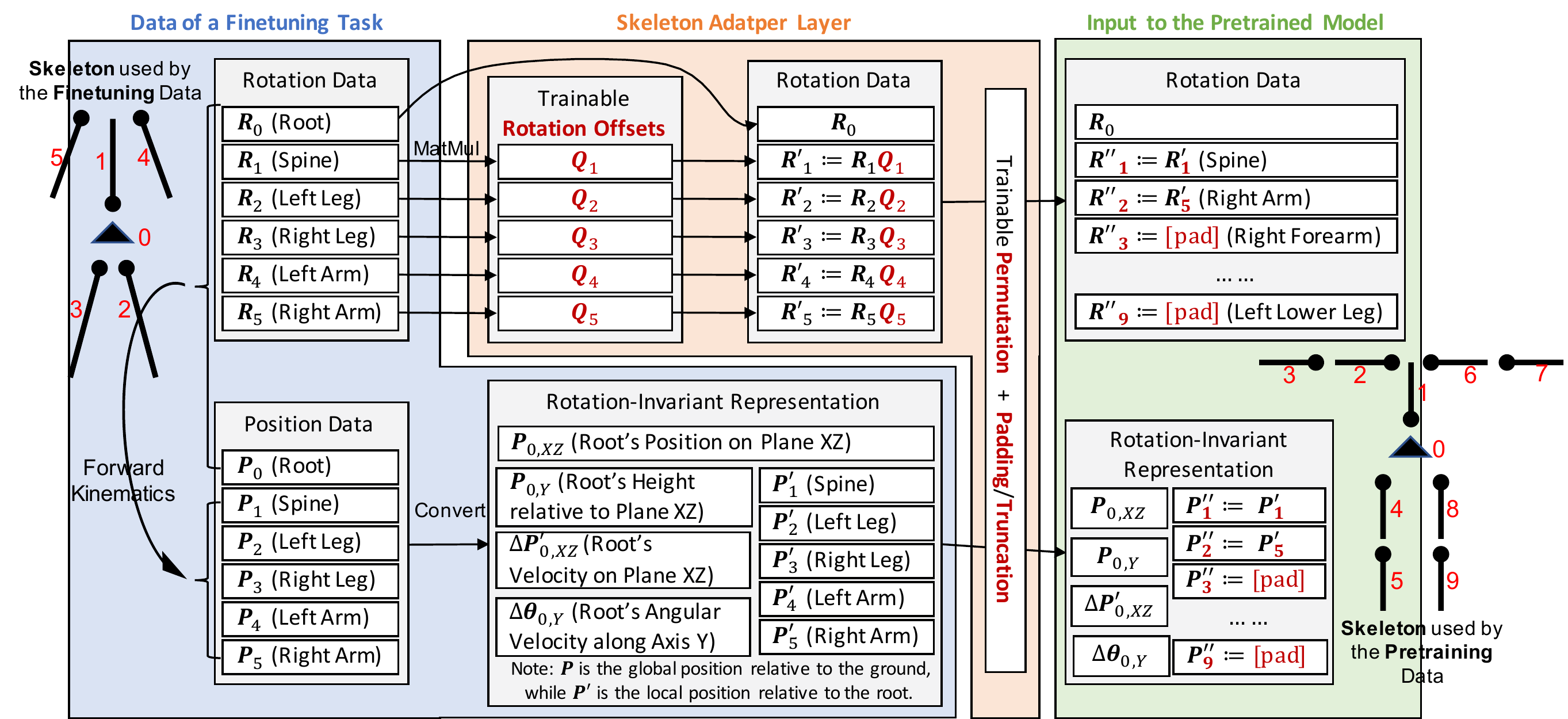}
        \caption{
            The data format and the skeleton adapter.
            The skeleton adapter is for bridging the gap between the skeleton used by fine-tuning (e.g., the left stick figure) and the skeleton used by pretraining (e.g., the right stick figure).
            Skeletons can differ in their default poses, number of joints, and indices of the joints.
        }
        \label{fig:main-skeleton}
    \end{subfigure}
    \caption{MoFusion, the proposed diffusion-based framework.
    It is pretrained on diverse motion datasets where there can be text control signals, music control signals, or no control signal, etc.
    The skeleton adapter is for fine-tuning only, and not included during pretraining.
    }
\end{figure*}

%% file: method.tex
\section{Method}
\label{sec:method}

We introduce the diffusion pipeline in Subsection~\ref{subsec:diffusion-pipeline}, the architecture and pretraining of MoFusion in Subsection~\ref{subsec:architecture-and-pretraining}, the skeleton adapter in Subsection~ \ref{subsec:data-format-and-skeleton-adapter}, and the fine-tuning procedure in Subsection~\ref{subsec:fine-tuning-and-inference}.
Finally, we discuss MoFusion's zero-shot ability in Subsection~\ref{subsec:zero-shot-generalization}.

\subsection{Diffusion Pipeline}\label{subsec:diffusion-pipeline}

We assume that each datapoint is a tensor in $\mathbb{R}^{N\times D}$, representing a motion clip consisting of $N$ frames and $D$ float numbers per frame.
MoFusion follows the pipeline of the de-noising diffusion probabilistic model (DDPM)~\cite{ho-ddpm}.

\paragraph{Inference procedure.}
DDPM aims to learn a de-nosing neural network $\mathbf{f}_\theta: \mathbb{R}^{N\times D}\times\mathbb{R}\to\mathbb{R}^{N\times D}$ for transforming any white noise tensor $\mathbf{x}_T \in \mathbb{R}^{N\times D}$ sampled from $\mathcal{N}(\boldsymbol{0}, \mathbf{I})$ into a meaningful datapoint $\mathbf{x}_0 \in \mathbb{R}^{N\times D}$ iteratively.
Here $T$ is the number of iterations, which is a hyperparameter.
The inference result $\mathbf{x}_0$ will be a synthesized motion clip in our setting.
We follow the previous Transformer-based diffusion models~\cite{dalle2,diffusionlm} and use the $\mathbf{x}_0$-prediction variant of the iterative de-noising process, where the $t$-th step is executed as follows:
\begin{gather}
    \hat{\mathbf{x}}_0 = \mathbf{f}_\theta ( \mathbf{x}_t , t)
    \label{eq:step-predict}
    \\
    \mathbf{x}_{t-1} \sim q( \mathbf{x}_{t-1} \mid \mathbf{x}_t, \hat{\mathbf{x}}_0, t )
    \label{eq:analytic-estimate}
    \\
    t = T, T-1, T-2, \ldots, 1
\end{gather}
The posterior distribution $q( \mathbf{x}_{t-1} \mid \mathbf{x}_t, \mathbf{x}_0, t )$ gives an estimate of $\mathbf{x}_{t-1}$ when given $\mathbf{x}_{t}$ and $\mathbf{x}_{0}$.
Eq.~(\ref{eq:analytic-estimate}) at the $t$-th step uses the model prediction $\hat{\mathbf{x}}_0$ as a substitute for $\mathbf{x}_0$, which is unknown yet, to approximately compute the posterior distribution.
In other words, it uses $q( \mathbf{x}_{t-1} \mid \mathbf{x}_t, \hat{\mathbf{x}}_0, t )$ in place of $q( \mathbf{x}_{t-1} \mid \mathbf{x}_t, \mathbf{x}_0, t )$.
The exact form of $q( \mathbf{x}_{t-1} \mid \mathbf{x}_t, \mathbf{x}_0, t )$ can be analytically derived after making a few assumptions, which we state in the training procedure below.

\paragraph{Training procedure.}
The neural network $\mathbf{f}_\theta (\cdot)$ is trained by minimizing the following de-noising loss:
\begin{gather}
    \mathcal{L}_\theta =
    \mathbb{E}_{\mathbf{x}_0, t} \mathbb{E}_{\mathbf{x}_t} [ \| \mathbf{f}_\theta (\mathbf{x}_t, t ) - \mathbf{x}_0 \|^2 ]
    \label{eq:train}
\end{gather}
Here the corrupted input $\mathbf{x}_t$ for training is obtained by iteratively corrupting $\mathbf{x}_0$ as follows:
\begin{gather}
    \mathbf{x}_t \sim \mathcal{N}( \sqrt{\alpha_t}\cdot \mathbf{x}_{t-1}, (1-\alpha_t)\cdot \mathbf{I}), \; t=1,2,3,\ldots,T
\end{gather}
The hyper-parameters $\{\alpha_t\}_{t=1}^T$ are decided by a cosine schedule~\cite{ddpm-cosine}.
The iterative corruption process is equivalent to the following simplified implementation:
\begin{gather}
    \mathbf{x}_t =
    \sqrt{\prod_{s=1}^t \alpha_t} \cdot \mathbf{x}_0 + \sqrt{1 - \prod_{s=1}^t \alpha_t} \cdot \boldsymbol{\epsilon},
    \;
    \boldsymbol{\epsilon} \sim \mathcal{N}( \boldsymbol{0}, \mathbf{I})
\end{gather}
This corruption process is chosen to make the posterior distribution $q( \mathbf{x}_{t-1} \mid \mathbf{x}_t, \mathbf{x}_0, t )$ used by Eq.~(\ref{eq:analytic-estimate}) analytically derivable, the exact closed-form solution of which is a Gaussian distribution (see Eq.~(6) of the DDPM paper~\cite{ho-ddpm}).

\paragraph{Conditional Synthesis.}
The neural network $\mathbf{f}_\theta ( \cdot )$ is designed such that it can optionally take a control signal $\mathbf{c}$ as input to describe the desired result.
For example, the signal can be a text.
We can then replace $\mathbf{f}_\theta ( \mathbf{x}_t , t)$ with $\mathbf{f}_\theta ( \mathbf{x}_t , t, \mathbf{c})$ in Eq.~(\ref{eq:step-predict}) and Eq.~(\ref{eq:train}) to support conditional synthesis.

\subsection{Pretraining a Backbone}\label{subsec:architecture-and-pretraining}

\paragraph{Backbone.}
MoFusion constructs the de-nosing network $\mathbf{f}_\theta ( \cdot )$ based on the Transformer-based encoder-decoder architecture~\cite{attention}, outlined in Fig.~\ref{fig:main-transformer}.
The decoder is bidirectional, takes a noisy motion clip $\mathbf{x}_t$ as well as the noise strength $t$ (i.e., the index of the current diffusion step) as input, and predicts a de-noised motion clip $\hat{\mathbf{x}}_0$.
A motion clip of shape $\mathbb{R}^{N\times D}$ is treated as a sequence of $N$ tokens.
Each token is projected to or from $\mathbb{R}^{D'}$ via a feed-forward layer, where $D'$ is the hidden dimension of the attention blocks.
The noise strength $t$ is projected into $\mathbb{R}^{D'}$ via a learnable embedding table, and is then added onto each input token.
We use an off-the-shelf encoder to encode the control signal $\mathbf{c}$ if available, e.g., DistilBERT~\cite{DistilBERT} for encoding a text.
The decoder receives the encoded signal via cross attention.

\paragraph{Pretraining.}
The skeleton adapter depicted in Fig.~\ref{fig:main-transformer} is for fine-tuning only and not included during the pretraining stage.
Pretraining of the backbone is carried out on data combined from multiple sources, such as AMASS~\cite{AMASS}, LaFAN1~\cite{ubi-inbetween}, KIT~\cite{KIT}, BABEL~\cite{BABEL}, Mixamo~\cite{mixamo}, 100STYLE~\cite{mason2022realtime}, AIST++~\cite{aist-dance-db,li2021learn}, and ChoreoMaster~\cite{choreomaster2021}.
The datasets do not use the same skeletons, with differences in the number of joints, joint indices, bone lengths, and the default poses.
We thus retarget them to the SMPL-H~\cite{SMPL,MANO} skeleton used by AMASS, based on a few heuristic rules handcrafted after investigating the datasets, and use the retargeted data for pretraining.
The encoder-related modules are skipped if a training sample does not contain control signals.
For samples that do contain control signals such as texts or music, we drop the control signal with a probability of 0.5 during pretraining.
The encoder's parameters are frozen when using an off-the-shelf encoder.

\subsection{Data Format and Skeleton Adapter}\label{subsec:data-format-and-skeleton-adapter}

It is necessary to discuss the data representation before we can explain the rationale behind the skeleton adapter.

\paragraph{Data format.}
MoFusion preprocesses a motion clip of $N$ frames into a tensor $\mathbf{x}_0 \in \mathbb{R}^{N\times D}$.
As illustrated in Fig.~\ref{fig:main-skeleton}, each frame is represented as a $D$-dimensional vector composed of the rotations and positions of the joints.
The rotation parameters describe how much the joints should rotate starting from the default pose.
The rotation of the root joint (usually the hip) is specified in the global coordinate system and describes the body's orientation, while the other joints specify their rotations relative to their parent joints according to a skeleton (e.g., the stick figures shown in Fig.~\ref{fig:main-skeleton}).
We additionally convert each $3\times3$ rotation matrix to its 6D vector representation~\cite{6d-rep}, i.e., the concatenation of the first two columns of the matrix since the last column is merely their cross product.
The root joint's 3D position and all the joints' rotations, along with the bone lengths provided by the skeleton, then uniquely determine the 3D positions of all the other joints, which is known as \emph{forward kinematics}.
Despite them being redundant, the 3D positions of all the joints are included in $\mathbf{x}_0$, because they are required by some tasks, such as \emph{inverse kinematics}.
We convert the 3D positions into the rotation-invariant format~\cite{holden-16-rotinvar} and perform $Z$-normalization to ease learning.

\paragraph{Skeleton adapter.}
A motion clip representing the same action can lead to different values of $\mathbf{x}_0$ with different skeletons, e.g., with the left stick figure versus with the right stick figure in Fig.~\ref{fig:main-skeleton}.
We manually retarget the data to the same skeleton for pretraining.
However, the evaluation benchmark for a fine-tuning task can use a different skeleton.
It is thus worthwhile to explore whether it is possible to automatically bridge the gap without manual retargeting by introducing a learnable adapter.

The gap comes from three differences between the skeletons.
The first difference is about bone lengths, which is a minor issue since the data are from actors of roughly the average body shape, and is addressed by simply rescaling the skeletons such that they are of the same height.

The second difference lies in the default pose, aka., the reference pose.
The rotation parameters measure the angles between the current pose and the default pose.
As a result, the rotation parameter of joint $j$, expressed as a $3\times 3$ matrix $\mathbf{R}_j\in\mathbb{R}^{3\times 3}$, needs to be offset by an extra rotation matrix $\mathbf{Q}_j\in\mathbb{R}^{3\times 3}$ if the default pose changes.
Our skeleton adapter module thus learns $\mathbf{Q}_j$, converts $\mathbf{R}_j$ to $\mathbf{R}_j\mathbf{Q}_j$ before sending the input to the pretrained Transformer backbone, and converts the pretrained Transformer backbone's output $\mathbf{R}'_j$ to $\mathbf{R}'_j \mathbf{Q}_j^{-1}$.
The implementation of the adapter treats the 6D representation of $\mathbf{Q}_j$ as trainable parameters.

The third difference is regarding the number of joints as well as the indices of the joints.
Fortunately, it is common to use the hip joint as the root index, i.e., the $0$-th joint.
Furthermore, the root joint typically have three subtrees corresponding to the two legs and the spine.
And the spine have two or three subtrees, i.e., the two arms and optionally the head.
Our adapter therefore uses these cues to group the joints into five or six groups\textemdash the left leg, the right leg, the spine, the left arm, the right arm, and optionally the head\textemdash by investigating the number of subtrees and whether there is a pair of symmetric subtrees.
Assuming the pretraining skeleton and the fine-tuning skeleton have $n_p$ and $n_f$ joints in the same group, respectively, it is then trivial to establish the mapping of the joint indices if $n_p= n_f$.
If $n_p\neq n_f$, our adapter then introduces a trainable matrix $\mathbf{M}\in \mathbb{R}^{n_p\times n_f}$ to represent the mapping, which can be viewed as a generalized permutation matrix with truncation (if $n_f > n_p$) or padding (if $n_f < n_p$).
The adapter converts the joints' parameter $\mathbf{P}\in\mathbb{R}^{n_f\times d}$ to $\mathbf{M}\mathbf{P}$ before sending the input to the pretrained Transformer backbone, and converts the backbone's output $\mathbf{P}'\in\mathbb{R}^{n_p\times d}$ to $\mathbf{M}^{+}\mathbf{P}'$, where $\mathbf{M}^{+}$ is the pseudo inverse of $\mathbf{M}$.
Mapping $\mathbf{M}$ is initialized as the top-left block of the identity matrix.

Prior to fine-tuning the whole model using task-specific training objectives, we freeze the backbone and fine-tune only the adapter's parameters, ignoring control signals and using the unconditional diffusion loss in Eq.~(\ref{eq:train}).

\subsection{Fine-Tuning Tasks}\label{subsec:fine-tuning-and-inference}

MoFusion supports fine-tuning two types of conditional synthesis tasks.
One encodes a control signal via an encoder, while the other is motion completion where some part of the motion is known and serves as the control signal.

\paragraph{Text-to-motion and music-to-dance.}
It is straightforward to plug in an off-the-shelf encoder to incorporate control signals such as texts and music, as outlined in Fig.~\ref{fig:main-transformer}.
For text-to-motion, we follow the previous work~\cite{petrovich22temos}, use a frozen DistilBERT~\cite{DistilBERT} for encoding texts, and minimize the $\ell_1$ loss rather than the $\ell_2$ loss in Eq.~(\ref{eq:train}) on the KIT dataset~\cite{KIT} that uses the MMM skeleton~\cite{Mandery2016b}.
We also tune the weight of the loss that corresponds to the root's movement.
For music-to-dance, we encode music using the feature extraction pipeline provided by the previous work~\cite{li2021learn} and finetune the model on the AIST++ dataset~\cite{li2021learn}.

\paragraph{Motion in-betweening.}
The pretrained model can perform in-betweening similar to image in-painting without fine-tuning,
by changing Eq.~(\ref{eq:step-predict}) during inference to
\begin{gather}
    \hat{\mathbf{x}}_0 =
    (1- \mathbf{m}) \cdot \mathbf{f}_\theta ( \mathbf{x}_t , t)
    + \mathbf{m} \cdot \mathbf{x}_{\textnormal{known}}
    \label{eq:inpaint}
\end{gather}
where $\mathbf{x}_{\textnormal{known}} \in \mathbb{R}^{N\times D}$ contains the known frames, while $\mathbf{m}\in \{0,1\}^N$ is a binary mask that indicates whether the $i$-th frame is known (if $\mathbf{m}[i]=1$) or unknown (if $\mathbf{m}[i]=0$).

However, our experiment shows that this non-finetuned approach tends to have noticeable discontinuity on the boundary of the predicted intermediate frames (see Fig.~\ref{fig:delta-trick-baseline}).
And we find that fine-tuning, along with a post-processing trick, can alleviate discontinuity.
Let a training sample for fine-tuning be $[\mathbf{x}_0^{(a)}; \mathbf{x}_t^{(b)}; \mathbf{x}_0^{(c)}] \in \mathbb{R}^{(N_a+N_b+N_c)\times D} $, where
$\mathbf{x}_0^{(a)}\in\mathbb{R}^{N_a\times D}$ and  $\mathbf{x}_0^{(c)}\in\mathbb{R}^{N_c\times D}$ are the given keyframes while $\mathbf{x}_t^{(b)}\in\mathbb{R}^{N_a\times D}$ is a corrupted version of the target intermediate frames $\mathbf{x}_0^{(b)}$.
Modified from the pretraining loss in Eq.~(\ref{eq:train}), our fine-tuning objective minimizes the $\ell_1$ loss:
\begin{gather}
    \mathbb{E}[
    \|
    \mathbf{f}_\theta
    (
    [\mathbf{x}_0^{(a)}; \mathbf{x}_t^{(b)}; \mathbf{x}_0^{(c)}],
    [\boldsymbol{0}; \mathbf{t};  \boldsymbol{0}]
    ) -
    [\mathbf{x}_0^{(a)}; \mathbf{x}_0^{(b)}; \mathbf{x}_0^{(c)}]
    \|
    ]
    \label{eq:inb-loss}
\end{gather}
The model no longer takes a single scalar $t$ for the whole sample.
Rather, it receives a vector $[\boldsymbol{0}; \mathbf{t};  \boldsymbol{0}]\in\mathbb{R}^{N_a+N_b+N_c}$, i.e., the known (or unknown) tokens should be added with a step embedding corresponding to step $0$ (or $t$).
Fine-tuning is conducted on dataset LaFAN1~\cite{ubi-inbetween} using its skeleton.

We then propose \emph{delta in-painting} for replacing Eq.~(\ref{eq:step-predict}) to further avoid discontinuity, which improves upon Eq.~(\ref{eq:inpaint}):
\begin{gather}
    \tilde{\mathbf{x}}_0 =
    \mathbf{f}_\theta
    (
    [\mathbf{x}_0^{(a)}; \mathbf{x}_t^{(b)}; \mathbf{x}_0^{(c)}],
    [\boldsymbol{0}; \mathbf{t};  \boldsymbol{0}]
    )
    \\
    \boldsymbol{\Delta}_t =
    \mathtt{slerp}( \mathbf{x}_{\textnormal{known}}, \mathbf{m} )
    -
    \mathtt{slerp}( \tilde{\mathbf{x}}, \mathbf{m} )
    \\
    \hat{\mathbf{x}}_0 =
    \tilde{\mathbf{x}} + \boldsymbol{\Delta}_t
    \label{eq:delta-inpaint}
\end{gather}
Here $\mathtt{slerp}(\mathbf{x}, \mathbf{m})$ is defined such that $\mathtt{slerp}(\mathbf{x}, \mathbf{m})[i]$ is equal to $\mathbf{x}[i]$ if $\mathbf{m}[i] = 1$, or is equal to the Slerp interpolation between $\mathbf{x}[a]$ and $\mathbf{x}[b]$ if $\mathbf{m}[i] = 0$, where $a = \max\{j: j < i \land m_j = 1\}$ and $b = \min\{j: j > i \land m_j = 1\}$.
Empirically, $\boldsymbol{\Delta}_t$ is nearly constant across the frames, and explains the difference in terms of the amplitude of motion.

\subsection{Zero-Shot Generalization}\label{subsec:zero-shot-generalization}
We define \emph{zero-shot learning} as the ability to perform a task without explicitly training for that specific task.

\paragraph{Modifying a body part.}
The pretrained model can readily modify the movement of a body part in specified frames while retaining the content of the others.
It is achieved by using the same technique shown in Eq.~(\ref{eq:inpaint}), with a mask of shape $\{0,1\}^{N\times D}$ to indicate the part that requires re-generation.
Moreover, we can optionally use control signals such as texts to guide the generation, since the pretraining stage involves both unconditional and conditional synthesis.

\paragraph{Inverse kinematics.}
\emph{Inverse kinematics} is about finding the appropriate rotation parameters such that the result of \emph{forward kinematics} satisfies the specified constraints.
The constraints usually specify the desired positions of a few joints, and the solution is in general not unique.
It can again be implemented via the technique in Eq.~(\ref{eq:inpaint}) by masking out the unknown parameters.
The predicted parameters however might not satisfy the constraints precisely.
We hence further refine the predicted parameters by minimizing the $\ell_1$ loss incurred by the constraints via gradient descent.

\paragraph{Mixing control signals.}
Let $\mathbf{c}_c$ and $\mathbf{c}_f$ be two control signals describing the desired coarse-grained feature and fine-grained feature, respectively.
For example, $\mathbf{c}_c$ can be a text vaguely describing an action such as ``walking'', and $\mathbf{c}_f$ be hip-hop music for specifying the fine-grained style.
Our pretrained model is able to synthesize a motion clip relevant to both signals, by replacing Eq.~(\ref{eq:step-predict}) with
\begin{gather}
    \hat{\mathbf{x}}_0 =
    \begin{cases}
        \mathbf{f}_\theta ( \mathbf{x}_t , t, \mathbf{c}_c) &  \textnormal{with probability } p_t
        \\
        \mathbf{f}_\theta ( \mathbf{x}_t , t, \mathbf{c}_f) & \textnormal{with probability } 1-p_t
    \end{cases},
    \\
    p_t = \left(\frac{t}{T}\right)^{\gamma}
\end{gather}
It favors $\mathbf{c}_c$ at the beginning of the de-noising process to determine the coarse-grained outline, and favors $\mathbf{c}_f$ to finalize the details when $t$ is near $0$.
The hyperparameter $\gamma \ge 0$ is for controlling the relative importance of the two signals, where a larger $\gamma$ favors $\mathbf{c}_f$ more.
We name this trick \emph{alternating control}, which requires no training sample to simultaneously have two control signals.

%% file: results.tex
\section{Experiments}
\label{sec:experiments}

We measure MoFusion's performance on in-betweening in Subsection~\ref{subsec:motion-in-betwening} and text-to-motion in Subsection~\ref{subsec:text-to-motion}, and investigate its zero-shot learning ability in Subsection~\ref{subsec:zero-shot-generalization2}.
Analyses on the key designs are in Subsection~\ref{subsec:ablation-studies}.
Source code and checkpoints will be released before publication.

The decoder of MoFusion comprises $12$ layers and $16$ attention heads with a hidden size of $1,024$, which has $\approx 250$M parameters.
The number of diffusion steps is $T=200$.
We use AdamW~\cite{adamw} with a batch size of $512$, and anneal the learning rate from $10^{-3}$ to $10^{-6}$.
We leave out $512$ samples for validation, pretrain MoFusion for $2,000$ epochs, and pick the best checkpoint according to validation.
We sample a subsequence of $w$ frames from a motion clip for training, where $w$ is a window size sampled from $[8,512]$, after down-sampling the motion clip to $12.5$ frames per second (FPS), which is the frame rate used by some tasks such as text-to-motion.
However, the in-betweening task requires $30$ FPS.
We thus continue the pretraining procedure for $1,000$ extra epochs with $30$ FPS before fine-tuning MoFusion for in-betweening.
Pretraining uses eight Nvidia V100 GPUs and finishes in less than a day.

\begin{table*}
    \centering
    \begin{tabular}{@{}lccccccccc@{}}
        \toprule
        & \multicolumn{3}{c}{{\bf L2Q$\downarrow$}} & \multicolumn{3}{c}{{\bf L2P$\downarrow$}} & \multicolumn{3}{c}{{\bf NPSS$\downarrow$}} \\
        \cmidrule(l){2-4}
        \cmidrule(l){5-7}
        \cmidrule(l){8-10}
        {\bf Method}                                        & @5         & @15        & @30        & @5         & @15        & @30        & @5           & @15           & @30          \\
        \midrule
        Zero-Vel                                            & 0.56       & 1.10       & 1.51       & 1.52       & 3.69       & 6.60       & 0.0053       & 0.0522        & 0.2318       \\
        Interp                                              & 0.22       & 0.62       & 0.98       & 0.37       & 1.25       & 2.32       & 0.0023       & 0.0391        & 0.2013       \\
        \midrule
        Kaufmann \etal~\cite{MotionInfilling} (Convolution) & 0.49       & 0.60       & 0.78       & 0.84       & 1.07       & 1.53 & 0.0048 & 0.0345 & 0.1454 \\
        Harvey \etal~\cite{ubi-inbetween} (LSTM)            & 0.17       & 0.42       & 0.69       & 0.23       & 0.65       & 1.28       & 0.0020       & 0.0258 & 0.1328 \\
        Duan \etal~\cite{duan2021unify} (Transformer)       & 0.14       & 0.36       & 0.61       & 0.22       & 0.56       & 1.10       & 0.0016 & 0.0234 & {\bf 0.1222} \\
        \midrule
        MoFusion (Ours)                                     & {\bf 0.12} & {\bf 0.33} & {\bf 0.58} & {\bf 0.14} & {\bf 0.51} & {\bf 1.09} & {\bf 0.0014} & {\bf 0.02300} & 0.1250 \\
        \bottomrule
    \end{tabular}
    \caption{Performance comparison of motion in-betweening on the LaFAN1 benchmark. The downward arrow ``$\downarrow$'' means lower scores are better, while ``@$k$'' means there are $k$ intermediate frames to predict.}
    \label{tab:main-inbetween}
\end{table*}

\begin{table*}
    \centering
    \begin{tabular}{@{}lcccccccc@{}}
        \toprule
        & \multicolumn{4}{c}{{\bf Average Positional Error (APE) $\downarrow$}}
        & \multicolumn{4}{c}{{\bf Average Variance Error (AVE) $\downarrow$}}
        \\
        \cmidrule(l){2-5}
        \cmidrule(l){6-9}
        {\bf Method}                     & {\small root } & {\small trajectory } & {\small local } & {\small global } & {\small root } & {\small trajectory} & {\small local} & {\small global} \\
        \midrule
        Lin \etal~\cite{lin-vigil18}     & 1.966          & 1.956                & 0.105           & 1.969            & 0.790          & 0.789               & 0.007          & 0.791           \\
        JL2P~\cite{lang2pose}            & 1.622          & 1.616                & 0.097           & 1.630            & 0.669          & 0.669               & 0.006          & 0.672           \\
        Ghosh \etal~\cite{ghosh2021-t2m} & 1.291          & 1.242                & 0.206           & 1.294            & 0.564          & 0.548               & 0.024          & 0.563           \\
        \midrule
        TEMOS@1~\cite{petrovich22temos}  & 0.963          & 0.955                & 0.104           & 0.976            & 0.445          & 0.445               & {\bf 0.005 }         & 0.448               \\
        MoFusion@1                       & {\bf 0.907 }   & {\bf 0.898 }         & {\bf 0.103 }    & {\bf 0.924 }     & {\bf 0.399 }   & {\bf 0.399 }          & {\bf 0.005 }        & {\bf 0.403}          \\
        \midrule
        TEMOS@10~\cite{petrovich22temos} & 0.784          & 0.774                & 0.104           & 0.802            & 0.392          & 0.391               & {\bf 0.005}               & 0.395          \\
        MoFusion@10                      & {\bf 0.779 }   & {\bf 0.769 }         & {\bf 0.103 }    & {\bf 0.798 }     & {\bf 0.354 }   & {\bf 0.353 }          & {\bf 0.005 }        & {\bf 0.358}          \\
        \bottomrule
    \end{tabular}
    \caption{Performance comparison of text-to-motion synthesis on the KIT benchmark.
    TEMOS and MoFusion are stochastic models, while the rest are deterministic.
    The suffix ``@$k$'' means we generate $k$ random samples and use the best one for computing the metrics.
    }
    \label{tab:main-text2motion}
\end{table*}

\begin{figure}
    \centering
    \begin{subfigure}{\linewidth}
        \includegraphics[width=1.00\linewidth]{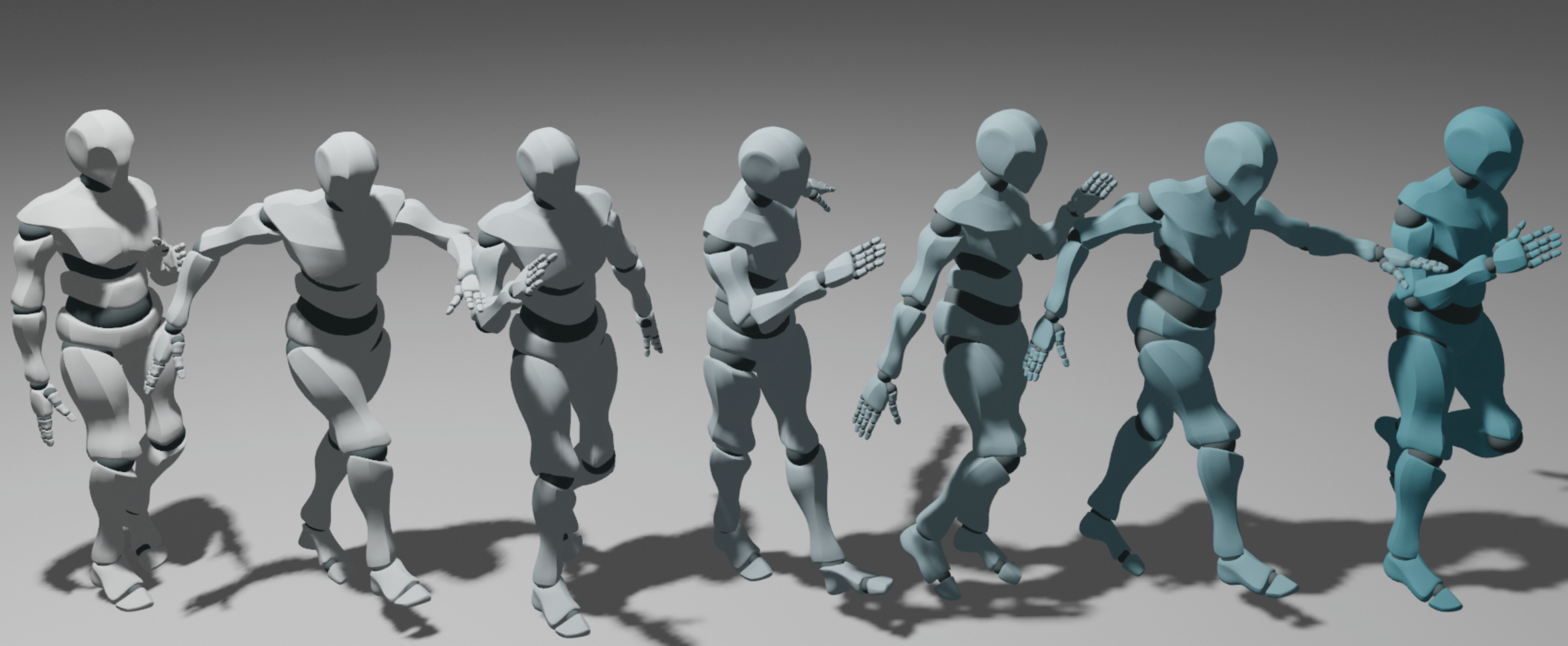}
        \caption{Step 1: Synthesize motion from scratch.}
        \label{fig:modify-part-1}
    \end{subfigure}
    \begin{subfigure}{\linewidth}
        \includegraphics[width=1.00\linewidth]{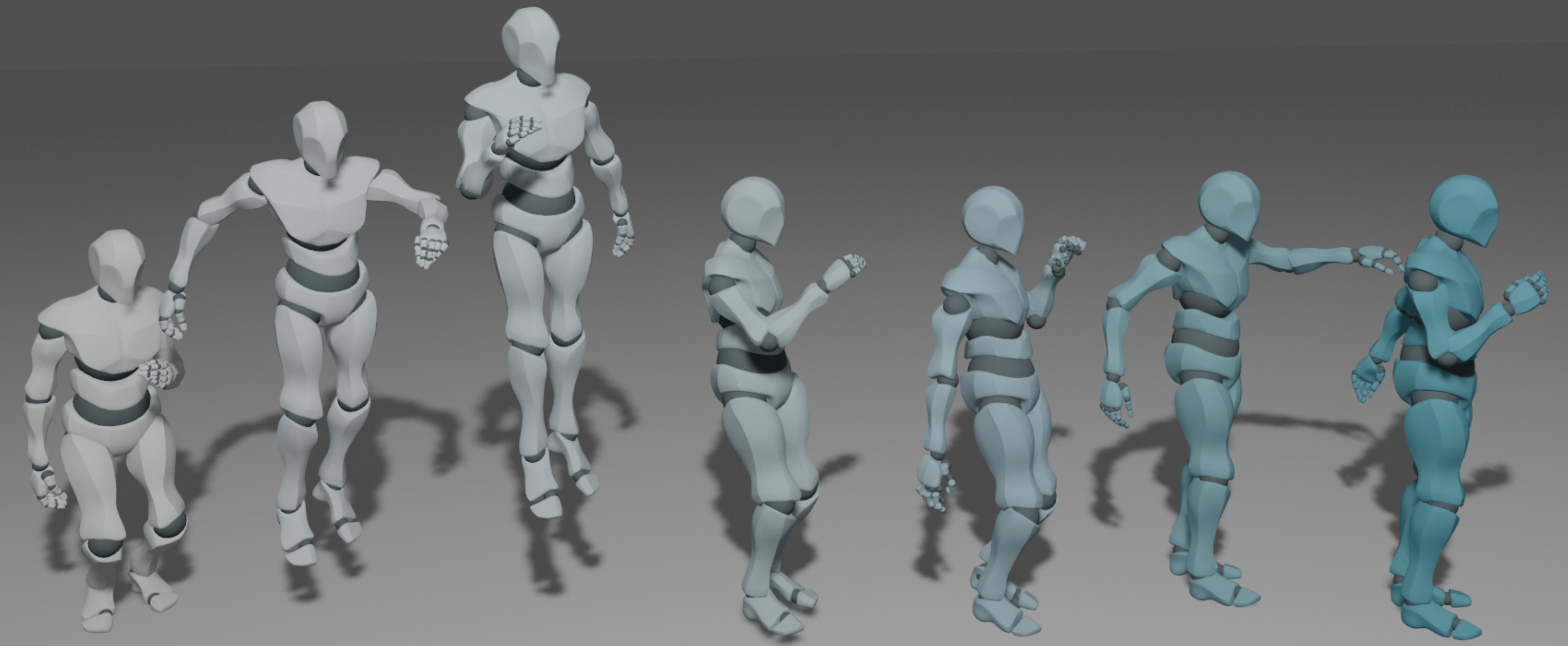}
        \caption{Step 2: Modify the lower body motion.}
        \label{fig:modify-part-2}
    \end{subfigure}
    \caption{
        MoFusion's zero-shot ability to modify the motion of body parts.
        The motion in Fig.~\ref{fig:modify-part-1} is synthesized from scratch conditioned on music labeled ``dance genre: House (120 BPM)''.
        The motion in Fig.~\ref{fig:modify-part-2} is modified from Fig.~\ref{fig:modify-part-1}, where the lower body motion is re-synthesized according to text ``jump with both legs'' while the motion of the spine and arms are kept unchanged.
    }
    \label{fig:modify-part}
\end{figure}

\begin{figure}
    \centering
    \begin{subfigure}{0.49\linewidth}
        \includegraphics[width=1.00\linewidth]{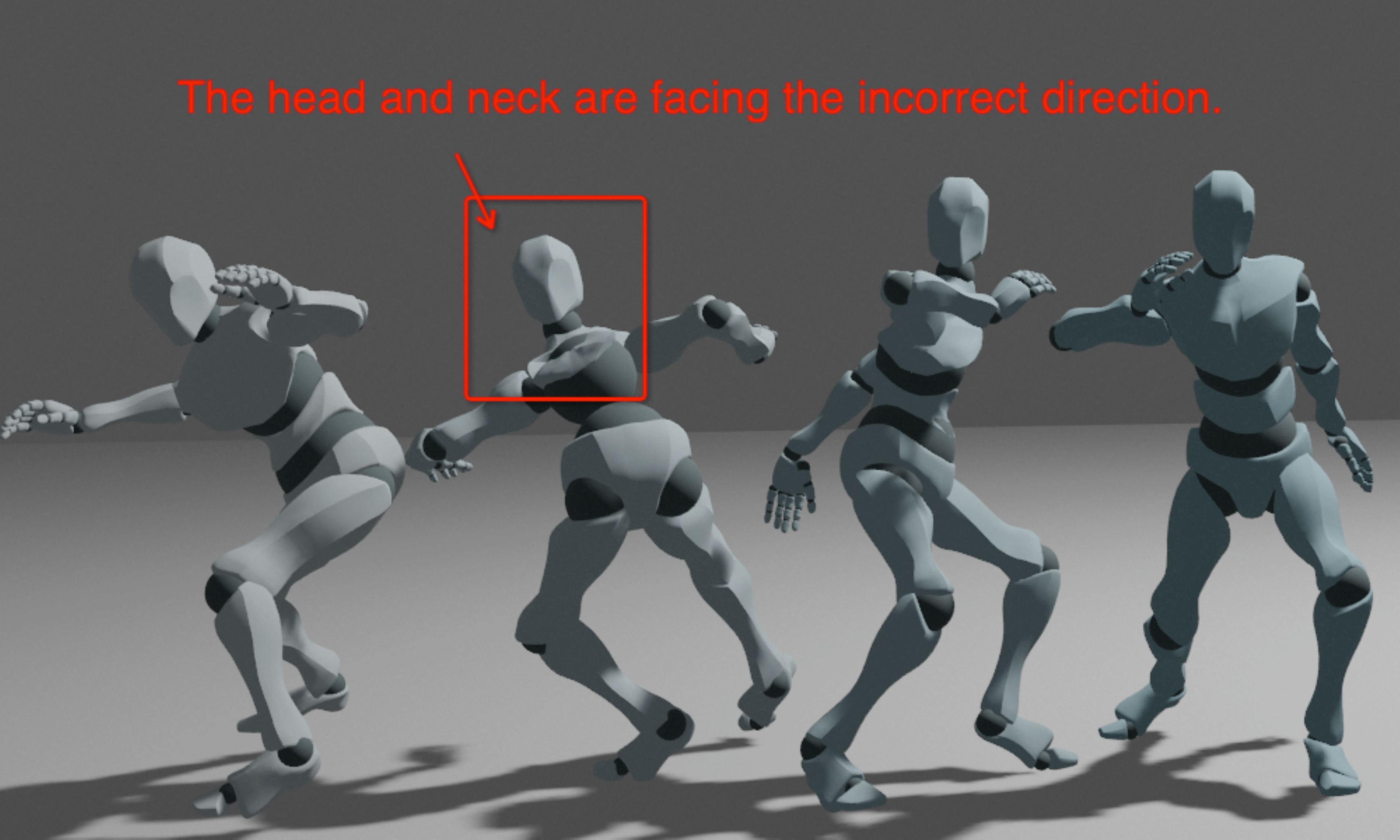}
        \caption{Rule-based baseline for IK.}
        \label{fig:inv-kine-1}
    \end{subfigure}
    \hfill
    \begin{subfigure}{0.49\linewidth}
        \includegraphics[width=1.00\linewidth]{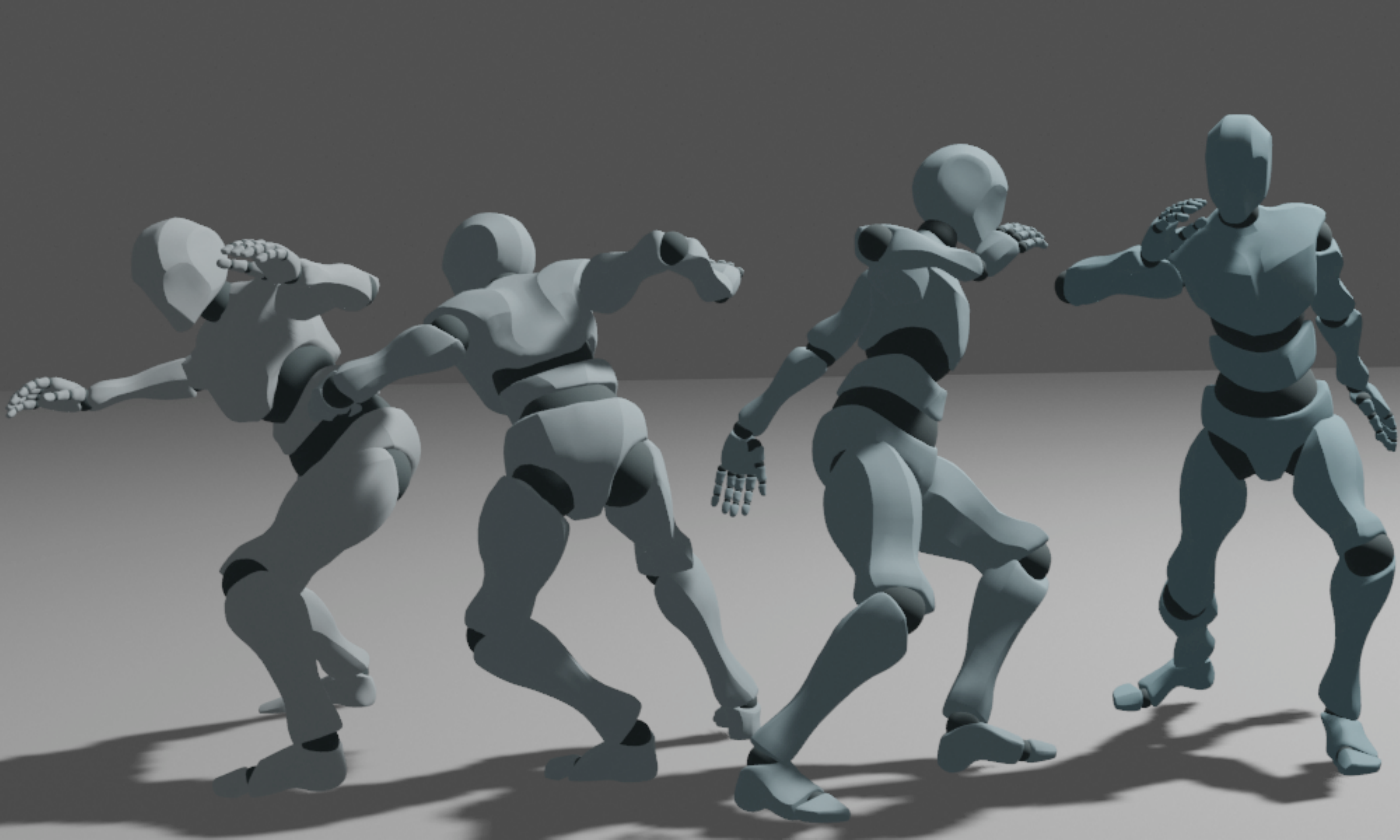}
        \caption{MoFusion for IK.}
        \label{fig:inv-kine-2}
    \end{subfigure}
    \caption{
        MoFusion's zero-shot ability to solve inverse kinematics (IK), where the task is to estimate the rotation parameters when given the positions of all the joints.
        The test case is a held-out sample titled ``360 degrees back leg sweep'' from Mixamo.
        The rule-based baseline fails to identify how much each bone should spin around itself, and thus leads to distorted mesh.
    }
    \label{fig:inv-kine}
\end{figure}

\begin{figure}
    \centering
    \begin{subfigure}{\linewidth}
        \includegraphics[width=1.00\linewidth]{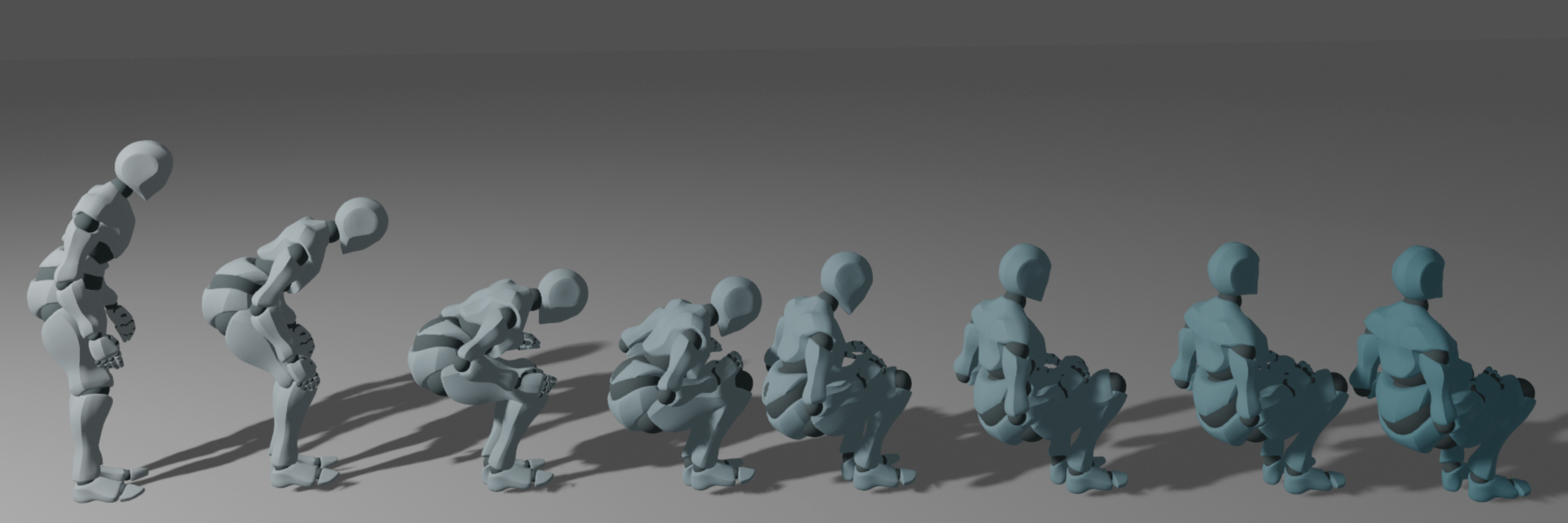}
        \caption{Text-to-motion synthesis.
        Equivalent to $\gamma = 0$}
        \label{fig:mixing-1}
    \end{subfigure}
    \begin{subfigure}{\linewidth}
        \includegraphics[width=1.00\linewidth]{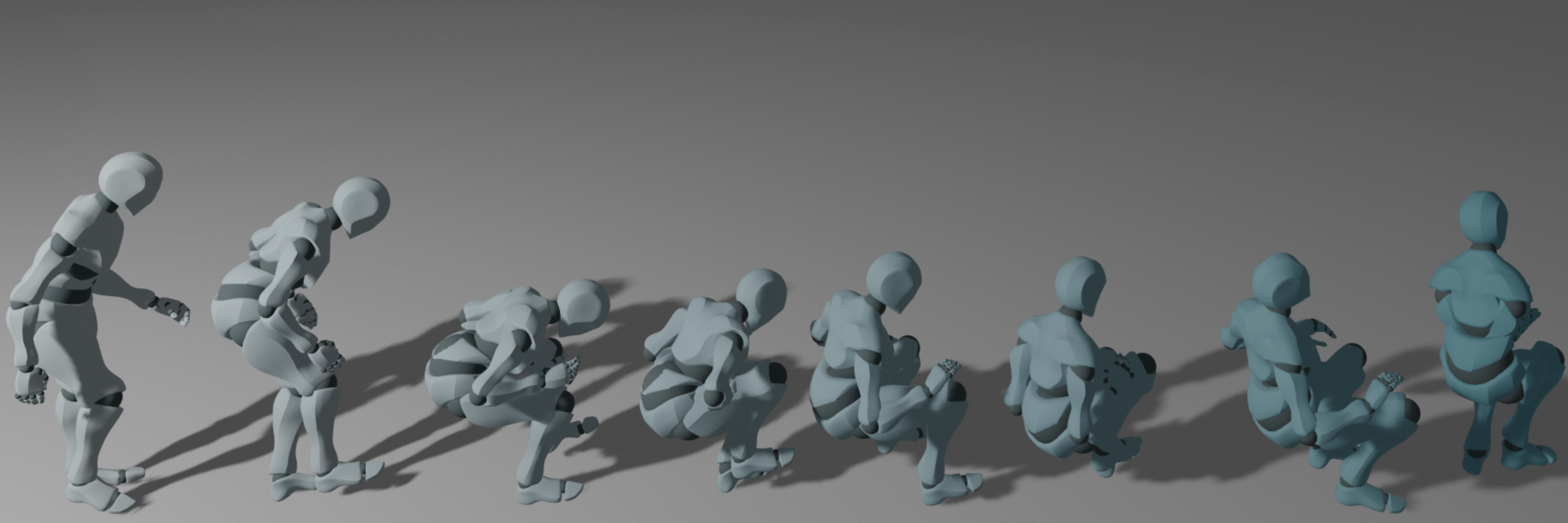}
        \caption{Mix the text and music signals.
        Hyperparameter $\gamma = 1.0$.}
        \label{fig:mixing-2}
    \end{subfigure}
    \begin{subfigure}{\linewidth}
        \includegraphics[width=1.00\linewidth]{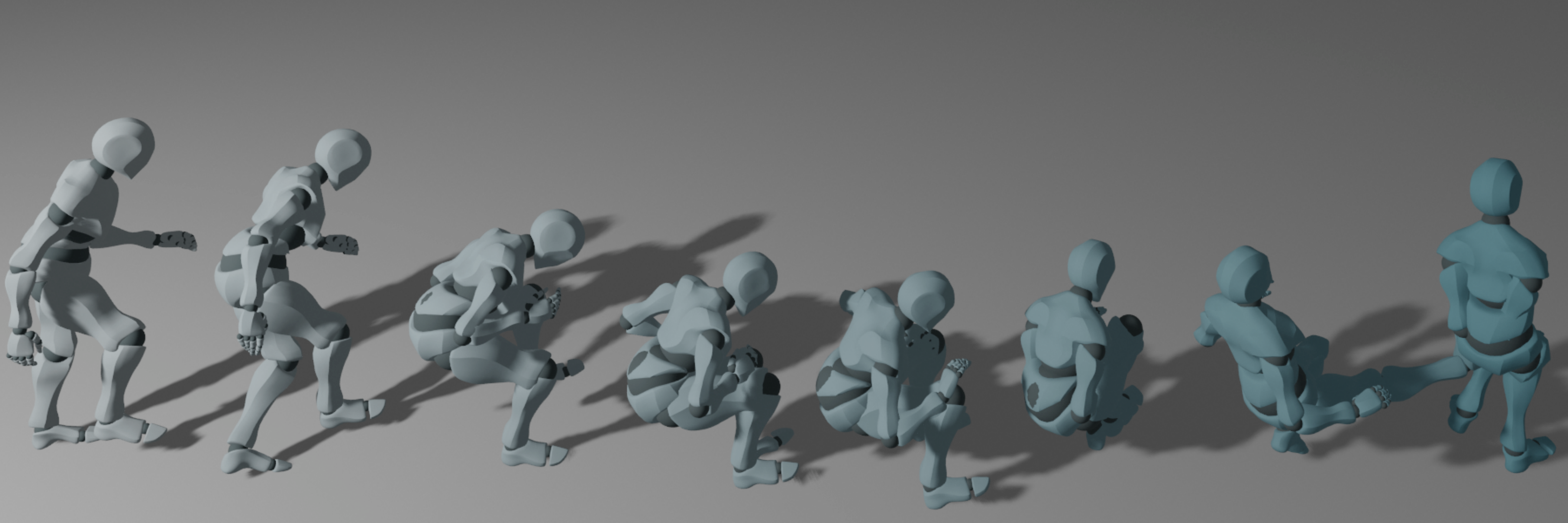}
        \caption{Mix the text and music signals.
        Hyperparameter $\gamma = 4.0$.}
        \label{fig:mixing-3}
    \end{subfigure}
    \begin{subfigure}{\linewidth}
        \includegraphics[width=1.00\linewidth]{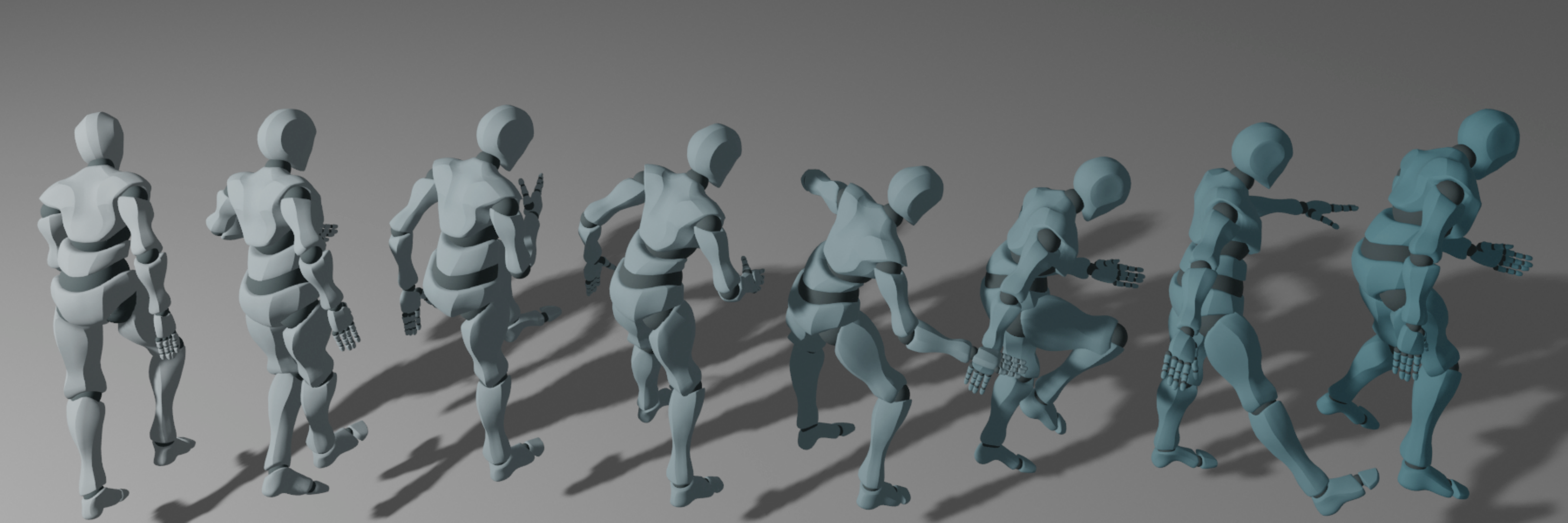}
        \caption{Music-to-dance synthesis.
        Equivalent to $\gamma \to +\infty$}
        \label{fig:mixing-4}
    \end{subfigure}
    \caption{
        MoFusion's zero-shot ability to mix control signals.
        The four samples use the same random seed, i.e., the same set of latent noises.
        The text signal is ``sit down on a chair'', while the music is labeled ``dance genre: Locking (130 BPM)''.
        Figs.~\ref{fig:mixing-2} and \ref{fig:mixing-3} preserve the semantics indicated by the text, but adjust the details to synchronize with the beats of the music, in different degrees.
    }
    \label{fig:mixing-result}
\end{figure}

\begin{table}
    \centering
    \begin{tabular}{@{}cclll@{}}
        \toprule
        & & \multicolumn{3}{c}{{\bf L2Q$\downarrow$}} \\
        \cmidrule(l){3-5}
        {\bf \#Parameters} & {\bf Pretrained?} & @5         & @15        & @30        \\
        \midrule
        50M                & No                & 0.14       & 0.36       & 0.61       \\
        250M               & No                & 0.16       & 0.38       & 0.64       \\
        \midrule
        50M                & Yes               & {\bf 0.12} & 0.35       & 0.61       \\
        250M               & Yes               & {\bf 0.12} & {\bf 0.33} & {\bf 0.58} \\
        \bottomrule
    \end{tabular}
    \caption{
        The impact of pretraining on model scaling.
        The other two metrics, L2P and NPSS, follow the same trend.
    }
    \label{tab:model-size}
\end{table}

\begin{table}
    \centering
    \begin{tabular}{@{}clll@{}}
        \toprule
        & \multicolumn{3}{c}{{\bf L2Q$\downarrow$}} \\
        \cmidrule(l){2-4}
        {\bf How to Adapt} & @5         & @15        & @30        \\
        \midrule
        Vanilla MLP        & 0.15       & 0.38       & 0.64       \\
        Manual Retargeting & {\bf 0.12} & {\bf 0.32} & {\bf 0.58} \\
        Skeleton Adapter   & {\bf 0.12} & 0.33       & {\bf 0.58} \\
        \bottomrule
    \end{tabular}
    \caption{Effectiveness of the proposed skeleton adapter.}
    \label{tab:adapter}
\end{table}

\begin{table}
    \centering
    \begin{tabular}{@{}clll@{}}
        \toprule
        & \multicolumn{3}{c}{{\bf L2Q$\downarrow$}} \\
        \cmidrule(l){2-4}
        {\bf Post-Processor}                          & @5         & @15        & @30        \\
        \midrule
        Vanilla In-Paint (Eq.~(\ref{eq:inpaint}))     & 0.14       & 0.34       & 0.59       \\
        RePaint~\cite{repaint}                        & 0.14       & 0.34       & {\bf 0.58} \\
        Delta In-Paint (Eq.~(\ref{eq:delta-inpaint})) & {\bf 0.12} & {\bf 0.33} & {\bf 0.58} \\
        \bottomrule
    \end{tabular}
    \caption{Effectiveness of our delta in-painting trick.
    The gain is especially significant when the number of intermediate frames is small, where the discontinuity issue affects the metrics more.
    }
    \label{tab:postprocess}
\end{table}

\begin{figure}
    \centering
    \begin{subfigure}{0.32\linewidth}
        \includegraphics[width=1.00\linewidth]{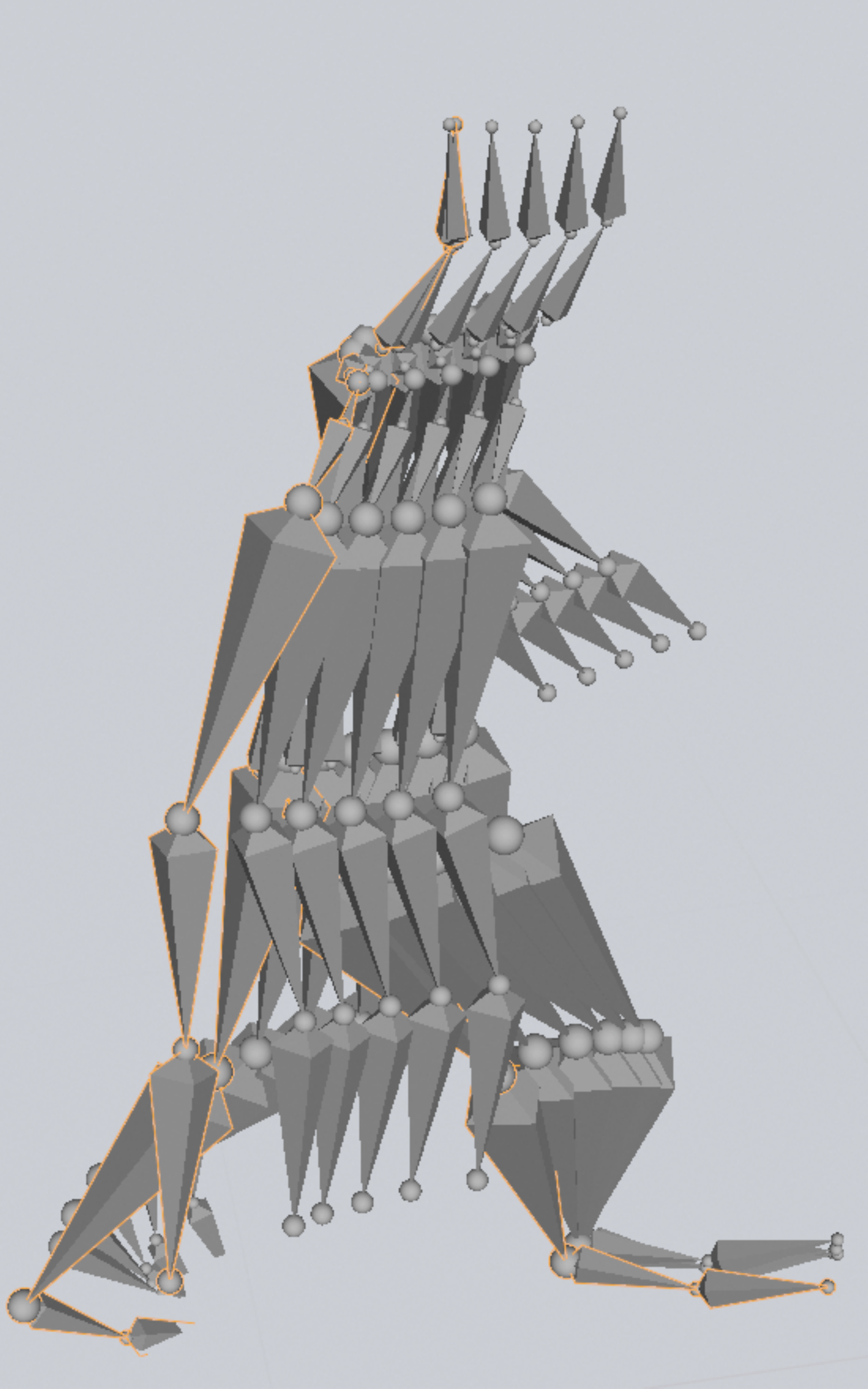}
        \caption{
            Vanilla in-painting (Eq.~(\ref{eq:inpaint})).
            No fine-tune.
        }
        \label{fig:delta-trick-baseline}
    \end{subfigure}
    \hfill
    \begin{subfigure}{0.32\linewidth}
        \includegraphics[width=1.00\linewidth]{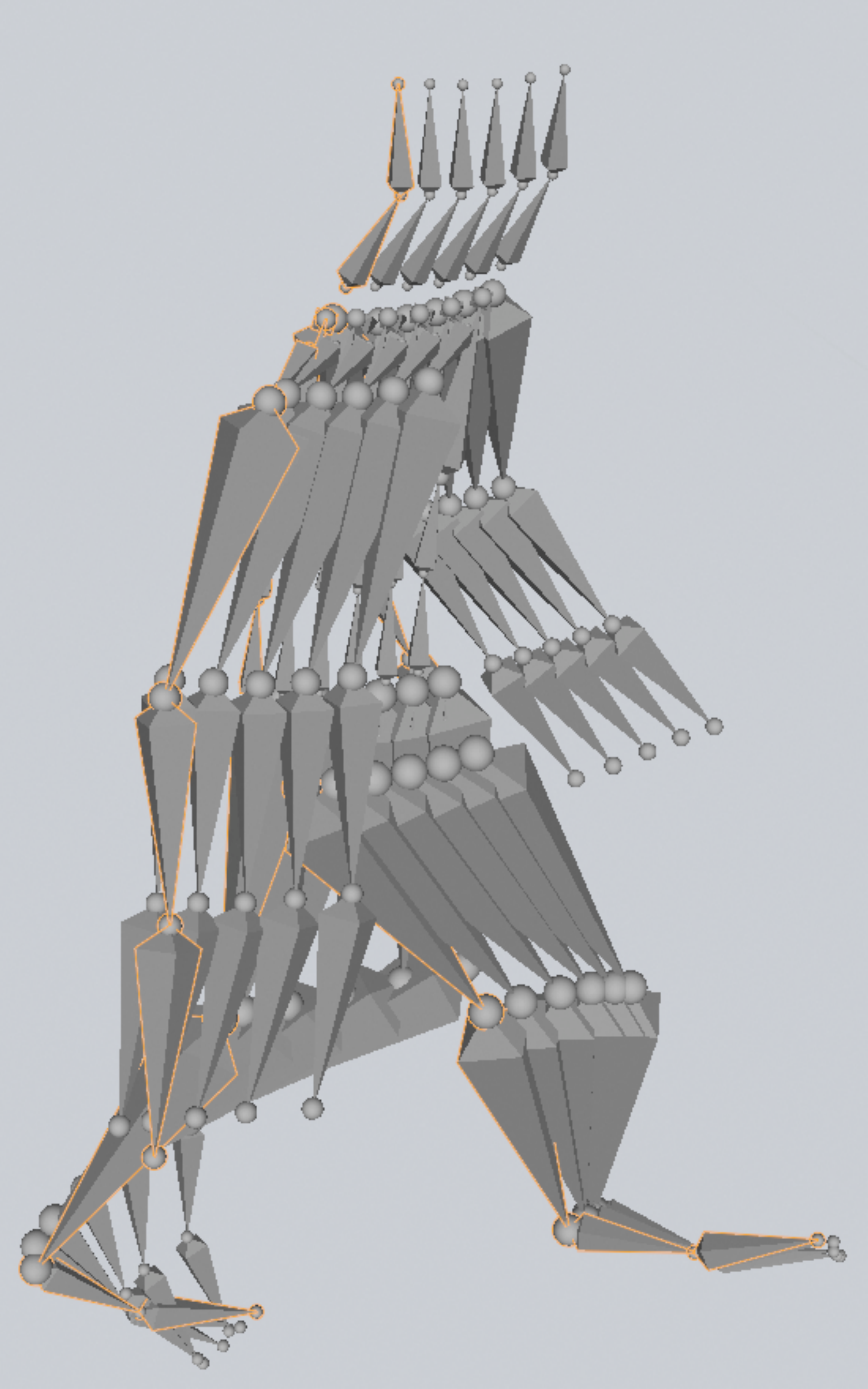}
        \caption{
            Vanilla in-painting (Eq.~(\ref{eq:inpaint})).
            Fine-tuned.
        }
        \label{fig:delta-trick-finetune}
    \end{subfigure}
    \hfill
    \begin{subfigure}{0.32\linewidth}
        \includegraphics[width=1.00\linewidth]{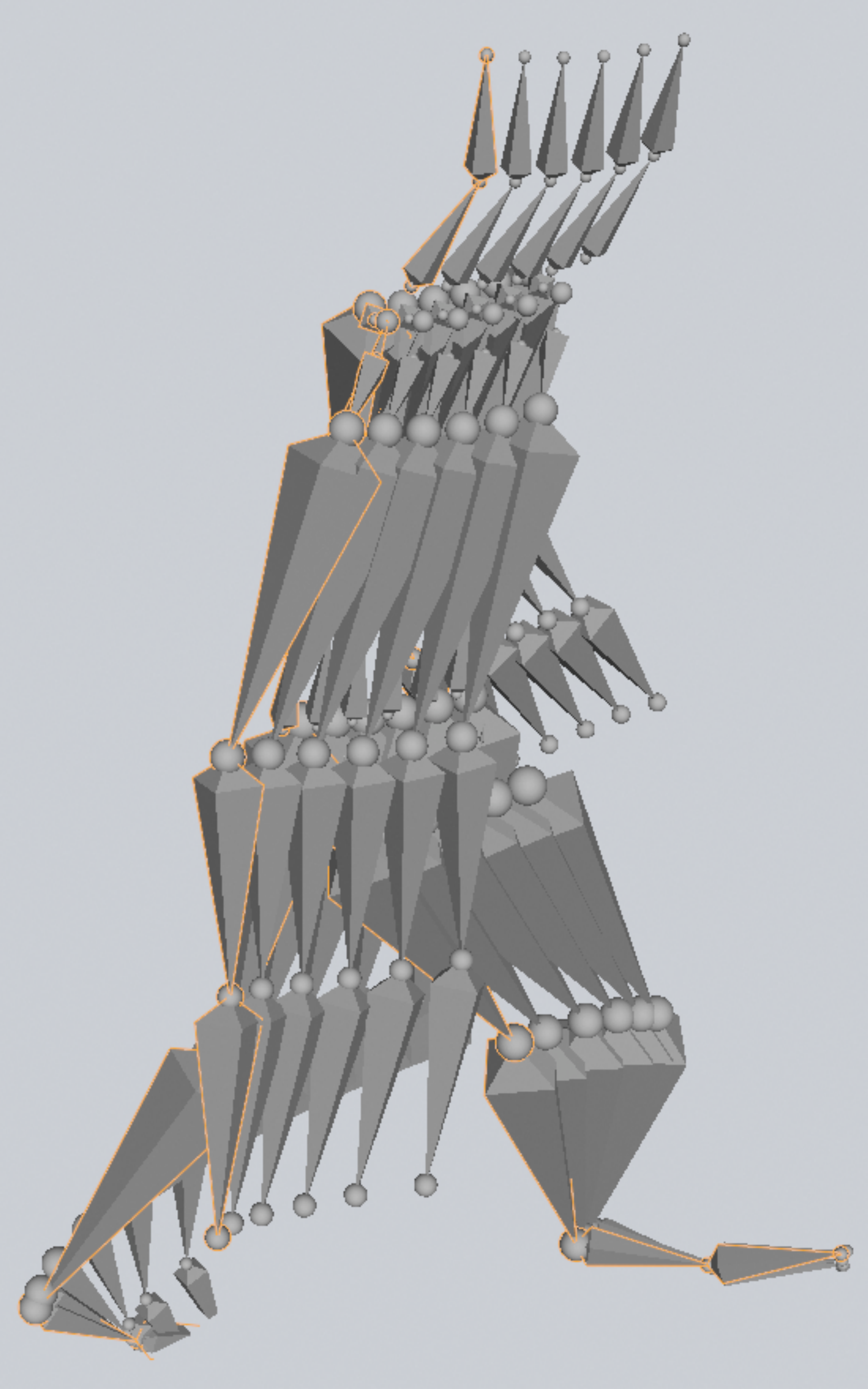}
        \caption{
            Delta in-painting (Eq.~(\ref{eq:delta-inpaint})).
            Fine-tuned.
        }
        \label{fig:delta-trick-full}
    \end{subfigure}
    \caption{
        Explicitly fine-tuning for motion in-betweening as in Eq.~\ref{eq:inb-loss}, along with the delta in-painting trick in Eq.~(\ref{eq:delta-inpaint}), can alleviate the discontinuity between the given keyframe (the yellow-highlighted frame) and the synthesized intermediate frames.
    }
    \label{fig:delta-trick}
\end{figure}

\subsection{Motion In-betwening}\label{subsec:motion-in-betwening}

We evaluate the performance for motion in-betweening on the LaFAN1 benchmark~\cite{ubi-inbetween}.
It provides ten keyframes at the beginning and one keyframe at the end of a motion clip, respectively.
The goal is to predict the $5/10/15$ intermediate frames.
We tune the skeleton adapter alone for five epochs, and then fully fine-tune MoFusion for $300$ epochs, with an annealed learning rate starting at $10^{-5}$ and a batch size of $64$ on a single V100 GPU.
In addition to the reconstruction loss in Eq.~(\ref{eq:inb-loss}), we also use the auxiliary losses introduced by Duan \etal~\cite{duan2021unify}.
As shown in in Tab.~\ref{tab:main-inbetween}, MoFusion outperforms the state of the art~\cite{duan2021unify} significantly.
More analyses are presented in Subsection~\ref{subsec:ablation-studies}.

\subsection{Text-to-Motion}\label{subsec:text-to-motion}

We follow the data sampling strategy and evaluation protocol established by TEMOS~\cite{petrovich22temos} when fine-tuning MoFusion for text-to-motion on the KIT dataset~\cite{KIT}.
AMASS~\cite{AMASS}, one of the data source for pretraining, has some overlaps with KIT.
We hence take extra care to ensure the validation and test cases of KIT are not used during pretraining.
We tune the skeleton adapter for five epochs and then finetune MoFusion for $500$ epochs with a learning rate starting at $10^{-4}$ and batch size of $32$ on a single V100 GPU.

The quantitative comparison is listed in Tab.~\ref{tab:main-text2motion}.
The APE metrics compare two motion clips frame by frame.
The AVE metrics first compute the variance of a motion clip across time, and then measure how similar two motion clips are in terms of their variances.
APE and AVE both have four variants: ``root'' considers only the root joint, ``trajectory'' is about the trajectory, ``local'' measures the poses after ignoring the root's orientation and movement, and ``global'' covers all the aforementioned aspects.

The improvement over the state of the art~\cite{petrovich22temos} is overall substantial, except on the ``local'' variant of the metrics.
A closer inspection of the KIT dataset reveals that most of the test cases are about ``walking'', where the diversity of the poses is very limited if the root orientation and trajectory are ignored.
Hence, we suspect the room for improvement on the ``local'' metrics is small.
The extra gain from generating multiple samples and picking the best corroborates the advantage of using a stochastic model for motion synthesis, as argued by TEMOS~\cite{petrovich22temos}.

\subsection{Zero-Shot Performance}\label{subsec:zero-shot-generalization2}

We call a setting \emph{zero-shot} if the model is not explicitly fine-tuned for a task before performing the task.

\paragraph{Modifying a body part.}
We investigate if the pretrained MoFusion can re-synthesize a specified part of a motion clip while keeping other parts intact, which is inspired by image diffusion models' in-painting ability~\cite{repaint}.
The example in Fig.~\ref{fig:modify-part} indicates that MoFusion does comes with this ability.

\paragraph{Inverse kinematics.}
As described in Subsection~\ref{subsec:data-format-and-skeleton-adapter}, the data format we use consists of both rotation parameters and position parameters.
Therefore, the problem of inverse kinematics can again be treated as ``in-painting'' the missing rotation parameters based on the position ones.
Fig.~\ref{fig:inv-kine} demonstrates MoFusion's ability to solve inverse kinematics.
The positions for all the joints are provided as input.
The rule-based baseline finds the rotation parameters by gradient descending from a randomly initialized starting point, while the MoFusion-based approach refines MoFusion's prediction via gradient descent.
We can see in Fig.~\ref{fig:inv-kine} that the rule-based baseline cannot correctly decide how much each bone should spin around itself, which leads to the distorted mesh.
This issue persists even after adding smoothness-related constraints or employing the twist-and-swing heuristic~\cite{hybrik}.
The MoFusion-based approach, on the contrary, produces natural-looking motion.

\paragraph{Mixing control signals.}
We use the proposed alternating control trick for zero-shot mixing two signals.
In Fig.~\ref{fig:mixing-result}, we mix a text as the coarse-grained signal with a piece of music as the fine-grained signal.
The mixed results are overall meaningful, and indicate that hyperparameter $\gamma$ can indeed control the relative importance of the two signals.

\subsection{Ablation Studies}\label{subsec:ablation-studies}

We now analyze the effectiveness of some key designs.

\paragraph{Pretraining and model scaling.}
In Tab.~\ref{tab:model-size}, we examine the relationship between pretraining and model scaling on the LaFAN1 benchmark~\cite{ubi-inbetween}.
The results show that a bigger model can perform much worse without pretraining, due to severe overfitting.
The previous works thus tend to use small models, i.e., $\ll$100M, while we show that a 250M big model can bring improvement once properly pretrained.

\paragraph{Skeleton adapter.}
In Tab.~\ref{tab:adapter}, we compared our skeleton adapter against two baselines: using a multilayer perceptron (MLP) in place of the skeleton adapter, or performing manual retargeting by hand.
The MLP has a single layer, as we observe no improvement with more layers.
Our skeleton adapter is competitive with manual retargeting, while the MLP baseline performs as if it were not pretrained.

\paragraph{Post-processing tricks.}
Fig.~\ref{fig:delta-trick} illustrates the discontinuity issue that happens when a non-finetuned diffusion model is used for in-betweening, as well as the improvement brought by our fine-tuning objective (i.e., Eq.~(\ref{eq:inb-loss})) and our delta in-painting trick (i.e., Eq.~(\ref{eq:delta-inpaint})).
The discontinuity issue in Fig.~\ref{fig:delta-trick-baseline} exists even with RePaint~\cite{repaint}, an improved trick proposed for image in-painting.
We also observe quantitative improvement brought by our methods, as listed in Tab.~\ref{tab:postprocess}.

%% file: related.tex
\section{Related Work}
\label{sec:related}

\paragraph{Diffusion models.}

Diffusion models~\cite{diffusion-sohl-dickstein15,song-diffusion} are deep generative models that model the data generating process as a Markov chain, according to which the datapoints diffuse through the latent space and arrive at a prior distribution.
They have recently gain tremendous success in the field of image synthesis~\cite{ho-ddpm,diffusionbeatgan}, especially text-to-image synthesis~\cite{glide,dalle2,ldm,imagen}.
Acceleration methods that reduce the de-nosing steps required have further improved the accessibility of these powerful models~\cite{ddim,liu2022pseudo}.
Techniques are proposed for conditional synthesis with diffusion models, e.g., classifier guidance~\cite{diffusionbeatgan} and classifier-free guidance~\cite{classifier-free}.

Concurrent to our work, some recent works employ diffusion models for text-to-motion synthesis~\cite{tevet2022human,zhang2022motiondiffuse,kim2022motiondiffuse}.
However, unlike our work, they remain task-specific, and do not explore multitask unification or pretraining.

\paragraph{Human motion synthesis.}
The field of motion synthesis have been gradually adopting neural networks due to their various advantages, such as being more efficient when scaling up the data size~\cite{learn-motion-match} compared to retrieval-based approaches~\cite{motiongraph-kovar02,motiongraph-lee02,motion-matching}.
Most of the recent works focus on the controllability of the synthesis process.
Some works specify the desired motion using control signals of other modalities, e.g., texts~\cite{ghosh2021-t2m,petrovich22temos,lin-vigil18,lang2pose}, music~\cite{li20-dance,siyao2022bailando,li2021learn,wenlinzhuang-dance,dance-revol}, and videos~\cite{aberman-retarget,aberman-transfer,VNect_SIGGRAPH2017,humanMotionKZFM19}.
And others focus on specifying some elements of the motion and predicting the rest, e.g., in-betweening~\cite{ubi-inbetween,MotionInfilling,duan2021unify}, inverse kinematics~\cite{oreshkin2022protores}, trajectory control~\cite{learn-motion-match}.
Most of them are for one task, and the few who do attempt unification consider tasks of limited scope, e.g., unifying only motion completion~\cite{duan2021unify}.
Our framework also includes an adapter to support varying skeletons, which is similar to deep motion retargeting~\cite{Villegas_2018_CVPR,aberman2020skeleton} while being more lightweight and not requiring dedicated training samples.

%% file: conclusion.tex
\section{Conclusion}
\label{sec:conclusion}

We have presented MoFusion, a pretrained diffusion-based framework for unifying motion synthesis.
We have demonstrated that unification not only improves the performance on conventional benchmarks, but also facilitates the emergence of zero-shot generalization and skill combination.
One future direction is to establish benchmarks for evaluating zero-/few-shot learning and skill combination.